\documentclass[lettersize,journal]{IEEEtran}
\usepackage[colorlinks,urlcolor=blue,linkcolor=blue,citecolor=blue]{hyperref}
  \usepackage{cite}
\usepackage{csquotes}
\usepackage{algorithm}
\usepackage{algpseudocode}
\usepackage{epsfig}
\usepackage{graphicx}
\usepackage{amsfonts}
\usepackage{comment}
\usepackage{color}
\usepackage{amssymb}
\usepackage{amsmath, mathtools}
\usepackage{amsthm}
\usepackage{etoolbox}
\usepackage{flushend}
\usepackage{babel,blindtext}
\usepackage{caption}
\usepackage{subcaption}
\usepackage{csquotes}
\usepackage{rotating}
\usepackage{textcomp}
\usepackage{footnote}
\usepackage{tablefootnote}
\usepackage[flushleft]{threeparttable}
\usepackage{paralist}
\usepackage{mdwlist}
\usepackage{url}
\usepackage{verbatim}
\usepackage{alltt}
\usepackage{pgfplots}
\usepgfplotslibrary{polar}
\pgfplotsset{compat=1.18} % use a recent version

\usepackage{multirow}
\usepackage{epstopdf}
\usepackage{lipsum}
\usepackage{float}
\usepackage{booktabs}
\usepackage{slashbox}
\usepackage[table]{xcolor}
\usepackage{array}
\usepackage{boldline}
\usepackage{caption,setspace}
\usepackage{stackengine}[2013-10-15]

\usepackage[T1]{fontenc}
\usepackage{frcursive}
\usepackage{calligra}
\usepackage{wedn}
\usepackage{bm}
\usepackage{aurical}
\usepackage{upgreek}

\definecolor{darkgreen}{RGB}{0,204,0}

\usepackage{cite}
\usepackage{adjustbox}
\usepackage{esint}
\usepackage{tipa}

\usepackage{tikz}
\usepackage{pgfmath}
\usepackage{pgfplots}
\pgfplotsset{compat=1.18}
\usepackage[dvipsnames]{xcolor}
\usetikzlibrary{calc}
\usetikzlibrary{matrix} % Load the matrix library

\definecolor{darkgreen}{RGB}{0,153,0}
\definecolor{darkred}{RGB}{192,0,0}

\usepackage[T1]{fontenc}

\newcolumntype{?}{!{\vrule width 1pt}}
\newcolumntype{+}{!{\vrule width 1.25pt}}

\def\hlineb#1{%
\noalign{\ifnum0=`}\fi\hrule \@height #1 %
\futurelet\reserved@a\@xhline}

\usepackage{cancel}

\definecolor{lightgray}{gray}{0.8}

\usepackage{stackengine}[2013-10-15]
\usepackage{ dsfont }
\usepackage[T1]{fontenc}
\usepackage{frcursive}
\usepackage{calligra}
\usepackage{wedn}
\usepackage{bm}
\usepackage{aurical}
\usepackage{ upgreek }

\usepackage{adjustbox}
\usepackage{xcolor}
\usepackage{makecell}
\usepackage{ mathrsfs }

\usepackage{pgfplots}
\pgfplotsset{compat=1.18}
\usepackage[svgnames,x11names,table]{xcolor}

\usepackage{csquotes}

\ifCLASSINFOpdf
\else
\fi
\begin{document}

\title{ProtoSiTex: Learning Semi-Interpretable Prototypes for Multi-label Text Classification}

\author{Utsav Kumar Nareti,~\IEEEmembership{Student Member,~IEEE}, 
Suraj Kumar,~\IEEEmembership{Student Member,~IEEE}, 
Soumya Pandey, 
Soumi Chattopadhyay,~\IEEEmembership{Senior Member,~IEEE}, 
Chandranath Adak,~\IEEEmembership{Senior Member,~IEEE},
Sankha Subhra Mullick
\thanks{
U.K. Nareti and C. Adak are with the Dept. of CSE, IIT Patna, India. 
S. Kumar, S. Pandey, and S. Chattopadhyay are with the Dept. of CSE, IIT Indore, India.  
S.S. Mullick is with Dolby Laboratories, India.
\emph{Corresponding author:} U.K. Nareti (email: utsav\_2221cs28@iitp.ac.in)
}
}

% \markboth{IEEE Transactions on Emerging Topics in Computational Intelligence}
% % {U. K. Nareti \MakeLowercase{\textit{et al.}}}
% {Anonymous \MakeLowercase{\textit{et al.}}}
% \author{Anonymous author
% \thanks{Anonymous author affiliation}
% }

% \markboth{IEEE Transactions on XX}
% {XXX}

\maketitle

\begin{abstract}
The rapid growth of user-generated text across digital platforms has intensified the need for interpretable models capable of fine-grained text classification and explanation.
% The surge in user-generated text has amplified the need for interpretable models that can provide fine-grained insights. 
Existing prototype-based models offer intuitive explanations but typically operate at coarse granularity (sentence or document level) and fail to address the multi-label nature of real-world text classification. We propose ProtoSiTex, a semi-interpretable framework designed for fine-grained multi-label text classification. ProtoSiTex employs a dual-phase alternate training strategy: an unsupervised prototype discovery phase that learns semantically coherent and diverse prototypes, and a supervised classification phase that maps these prototypes to class labels. A hierarchical loss function enforces consistency across subsentence, sentence, and document levels, enhancing interpretability and alignment. Unlike prior approaches, ProtoSiTex captures overlapping and conflicting semantics using adaptive prototypes and multi-head attention. We also introduce a benchmark dataset of hotel reviews annotated at the subsentence level with multiple labels. Experiments on this dataset and two public benchmarks (binary and multi-class) show that ProtoSiTex achieves state-of-the-art performance while delivering faithful, human-aligned explanations, establishing it as a robust solution for semi-interpretable multi-label text classification. 
\end{abstract}


\section{Introduction}\label{sec:intro}
In recent years, user-generated text across domains such as e-commerce, hospitality, and social media has grown rapidly \cite{review_helpfulness}, providing abundant but highly unstructured linguistic data \cite{online_review, online_review_1, online_review_2}. Extracting meaningful information from such textual content is essential for a wide range of NLP applications, and transformer-based language models (LMs) \cite{devlin-etal-2019-bert} have achieved state-of-the-art performance in tasks including sentiment analysis \cite{sentiment_survey}, question answering \cite{QA_survey}, and text classification \cite{TC_survey}. Despite their effectiveness, these models operate as opaque black boxes, offering limited interpretability, an increasingly critical requirement for trust, fairness, and regulatory compliance \cite{interpretability_1, interpretability_2}.
To mitigate the limitations of black-box behavior, post-hoc explanation techniques such as LIME \cite{LIME} and SHAP \cite{SHAP} have been widely adopted. However, these methods often struggle with faithfulness and consistency, producing approximations that may be difficult to interpret or unreliable for decision support \cite{interpretability_3, interpretability_4}. These challenges have motivated growing interest in inherently interpretable models that provide transparency by design \cite{interpretability_5}.

Prototype-Based Networks (PBNs) follow the paradigm of inherent interpretability, mimicking human reasoning via similarity to class-representative prototypes \cite{cognitive}. While effective in vision \cite{proto_img1, proto_img2} and recently adapted to NLP \cite{10.5555/3648699.3648963-ProtoryNet, wen-etal-2025-gaprotonet}, most PBNs are limited to single-label, sentence- or document-level tasks. This limits their use in real-world, multi-label classification scenarios, e.g., “\textit{The view was breathtaking, but the staff was rude}” expresses opposing sentiments, challenging coarse-grained prototype models \cite{sourati-etal-2024-robust}.

Efforts to improve interpretability in text classification often come at the expense of predictive strength. Transformer-based models capture rich semantic structures and typically outperform simpler methods in benchmarks \cite{10.1145/3495162-text-class-survey}, yet their lack of transparency limits trust in sensitive applications. Prototype-driven approaches \cite{protolens, wen-etal-2025-gaprotonet} provide clearer, case-based explanations that are easier for users to follow. However, they may sacrifice raw accuracy or struggle to scale to complex domains. This creates a persistent issue: optimizing solely for accuracy risks overlooking fairness and accountability, while prioritizing interpretability may reduce competitiveness in performance-driven evaluations. The issue becomes even more pronounced in multi-label classification \cite{pontiki2016semeval, schouten2015survey}, where models must balance fine-grained predictive capability with explanations that remain faithful and comprehensible.

To address these limitations, we propose {ProtoSiTex}, a semi-interpretable prototype-based framework for fine-grained multi-label classification. ProtoSiTex effectively handles overlapping sentiments and multi-label scenarios via three key innovations: 
(a) adaptive prototype learning using multi-head attention for subsentence-level semantic alignment, 
(b) a dual-phase alternate training strategy that combines unsupervised prototype discovery with supervised classification to improve generalization and reduce overfitting, and 
(c) a hierarchical loss function that enforces consistency across subsentence, sentence, and document levels for structured and faithful predictions. We use the term semi-interpretable to denote models that provide faithful, prototype-based explanations for their predictions while still relying on latent neural mechanisms (e.g., attention, embeddings, hierarchical aggregation) that are not inherently transparent. Our key contributions are summarized as follows:

\textit{\textbf{(i) A semi-interpretable prototype-based framework for fine-grained multi-label text classification:}} 
We propose ProtoSiTex, a prototype-driven framework that performs fine-grained multi-label text classification while providing intrinsic interpretability. Unlike prior prototype-based methods operating at sentence- or document-level granularity, ProtoSiTex learns adaptive prototypes aligned with subsentence-level semantic patterns. This enables effective disentanglement of overlapping and conflicting semantics through localized prototype specialization. Furthermore, the integration of multi-head attention facilitates context-aware prototype assignment and improves semantic alignment between textual evidence and labels.

\textit{\textbf{(ii) Dual-phase alternate optimization and hierarchical supervision for structured semantic learning:}} 
We introduce a dual-phase alternate training strategy that decouples unsupervised prototype discovery from supervised classification. The clustering phase promotes semantically coherent, diverse, and well-separated prototype representations, improving latent-space geometry and reducing prototype collapse. The classification phase employs hierarchical supervision across subsentence, sentence, and document levels, enforcing semantic consistency and structured label propagation. This formulation improves generalization, stabilizes prototype evolution, and enables reliable multi-label prediction with faithful prototype-grounded explanations.

\textit{\textbf{(iii) Prototype-grounded interpretability with adaptive semantic aggregation:}} 
ProtoSiTex establishes explicit alignment between subsentence-level evidence, adaptive prototypes, and document-level predictions through hierarchical differentiable aggregation. The aggregation mechanism acts as a continuous relaxation of set-union style label propagation, enabling end-to-end optimization while preserving semantic consistency across subsentence, sentence, and document levels. This facilitates traceable prototype-grounded explanations for multi-label predictions while maintaining competitive predictive performance.

We introduce a hotel reviews dataset (HR) annotated at the subsentence level with multi-label aspect annotations and explanation spans, supporting interpretability-grounded evaluation. Extensive experiments on HR, IMDb, and TweetEval demonstrate that ProtoSiTex consistently outperforms existing interpretable and prototype-based baselines while achieving competitive performance with strong black-box transformer and LLM-based models across binary, multi-class, and multi-label classification tasks.

The rest of the paper is organized as follows: 
Section \ref{sec:related_work} briefly reviews the literature, 
Section \ref{sec:methodology} presents our proposed approach, 
Section \ref{sec:experiments} details the experiments, and 
Section \ref{sec:conclusion} concludes with directions for future work.

\section{Related Work}\label{sec:related_work}
% This section briefly surveys related work and positions our study within the literature.

\textbf{\emph{Interpretable Text Classification:}}
Deep models like CNNs, RNNs, and transformers have achieved strong performance across NLP tasks \cite{10.1145/3495162-text-class-survey}, but their opaque nature hinders transparency. Post-hoc explanation methods, such as LIME \cite{LIME} and SHAP \cite{SHAP}, aim to interpret predictions by approximating local feature importance. However, these techniques operate independently of model training and often produce inconsistent or non-faithful rationales \cite{cesarini2024explainable-posthoc-limitation}. This has spurred interest in interpretable models that learn to classify and explain simultaneously.

\textbf{\emph{Prototype-Based Networks (PBNs):}} 
PBNs classify instances based on similarity to learned prototype vectors, offering human-aligned explanations via exemplar-based reasoning \cite{snell2017prototypical_CV}. Early works such as ProSeNet \cite{10.1145/3292500.3330908-ProsNet} and ClassVector \cite{10.1145/3318299.3318307-ClassVector} introduced prototype learning for document-level, single-label tasks with limited interpretability. Subsequent methods like ProtoryNet \cite{10.5555/3648699.3648963-ProtoryNet} and ProtoLens \cite{protolens} learned finer-grained prototypes at the sentence or subsentence level. GAProtoNet \cite{wen-etal-2025-gaprotonet} integrated graph-attention with prototypes for relational reasoning. Yet, most PBNs focus on single-label classification, lacking the granularity and compositionality required for multi-label classification.

% \emph{Multi-Label Aspect Classification:}
% \textcolor{red}{Aspect-level sentiment analysis and opinion mining \cite{pontiki2016semeval} require identifying and labeling multiple overlapping aspects within a document, often expressed at varying granularities \cite{hu2004mining, schouten2015survey}}. Existing methods either treat this as a flat multi-label classification task or rely on black-box architectures with no interpretability guarantees. Furthermore, few datasets provide sub-sentence-level multi-label annotations, limiting empirical benchmarking.

\textbf{{\emph{Multi-Label Text Classification (MLTC):}}}
MLTC involves assigning multiple labels to a document, making it more challenging than single-label tasks due to inherent label dependencies and data imbalance \cite{multi_label_tpami}. Early approaches relied primarily on problem transformation methods, such as binary relevance, label power set, and classifier chains, as well as algorithm adaptation methods, including multi-label SVMs and KNNs \cite{multi_label_methods}. Models based on CNNs and RNNs improved text representation learning \cite{multi_label_RNN, multi_label_cnn}, while GCNs were introduced to explicitly capture label correlations \cite{multi_label_gcn}. More recently, transformer-based architectures, including BERT \cite{devlin-etal-2019-bert}
%, X-MLClass \cite{XMLClass}, and LD-VAE \cite{LD-VAE},
have emerged as strong baselines. Despite these advances, existing methods either treat MLTC as a flat classification problem or rely on complex black-box architectures with limited interpretability. Furthermore, there is hardly any dataset that provides subsentence-level multi-label annotations, limiting empirical benchmarking.

\textbf{\emph{Positioning of Our Work:}}
{ProtoSiTex} overcomes these limitations by unifying prototype-based interpretability with fine-grained multi-label classification. Unlike prior works, it learns adaptive prototypes aligned with subsentence semantics and leverages a dual-phase learning strategy, combining unsupervised prototype discovery and supervised hierarchical classification. This design enables structured predictions from subsentence to document level while providing reliable, human-understandable explanations.

\section{Proposed Methodology}\label{sec:methodology}
% In this section, we present ProtoSiTex, our proposed method for semi-interpretable text classification. 
\begin{figure*}[!hbt]
    \centering
    \includegraphics[width=0.98\linewidth]{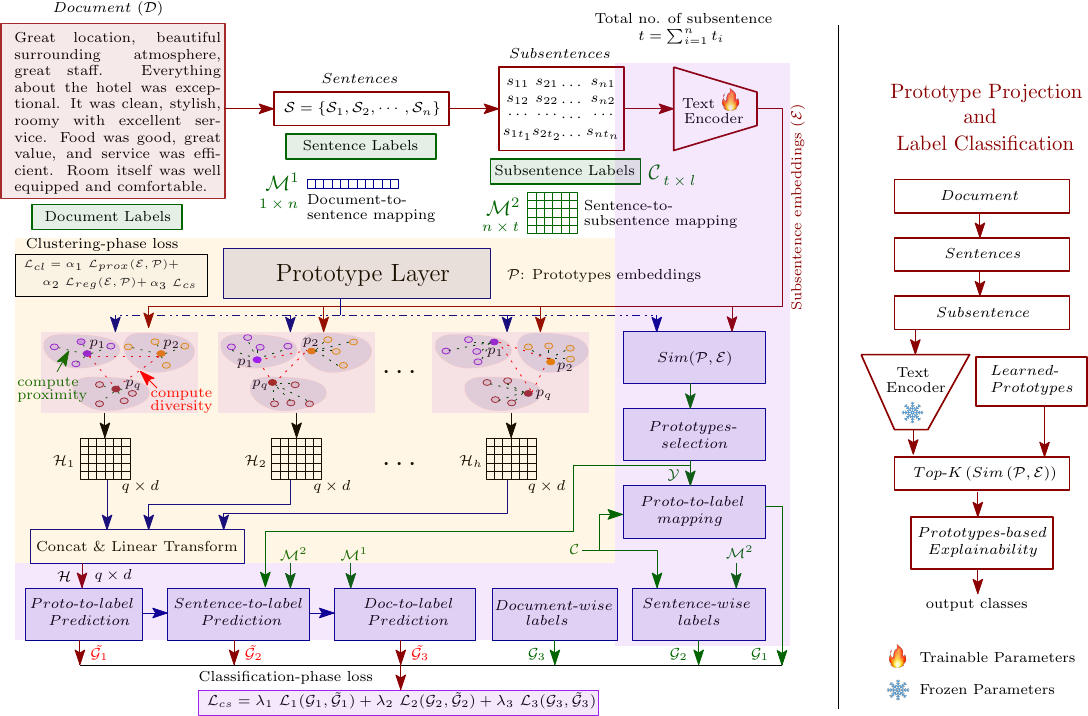}
    \caption{Proposed solution architecture: ProtoSiTex}
    \label{fig:arch}
\end{figure*}

% \textit{Problem formulation}: 
\subsection{Problem Formulation}

Given a document $\mathcal{D}$ composed of multiple sentences, our goal is to predict its associated class(es), accommodating both single-label and multi-label classification. ProtoSiTex performs document-level prediction while highlighting the specific content segments responsible for each label. 

% \noindent
% \textit{Preprocessing}: In our framework, subsentence annotations, though optional, enhance fine-grained interpretability in multi-label tasks. Since manual labeling is costly, we use a few-shot LLM-based annotation engine~\cite{ARTICLE_AAAI_2025} to infer them from document-level labels automatically.

Each document \(\mathcal{D}\) comprises \(n\) sentences, represented as {{\(\mathcal{S} = \{\mathcal{S}_1, \mathcal{S}_2, \ldots, \mathcal{S}_n\}\)}}. Each sentence \(\mathcal{S}_i\) is further segmented into \(t_i\) subsentences, defined as \(\mathcal{S}_i = \{s_{i1}, s_{i2}, \ldots, s_{it_i}\}\), where punctuation marks such as commas and semicolons are used as segmentation cues. The total number of subsentences in a document is denoted by \(t = \sum_{i=1}^{n} t_i\).
To assign fine-grained class labels, we apply the LLM-based engine to generate multi-label annotations at the subsentence level. These annotations are encoded in a binary label matrix \(\mathcal{C} \in \{0,1\}^{t \times l}\), where \(l\) is the total number of classes. A matrix entry \(c_{ij} = 1\) indicates that the \(i^\text{th}\) subsentence is associated with the \(j^\text{th}\) label.
For hierarchical supervision, we aggregate subsentence labels to form sentence-level labels, while document-level labels are directly provided. This bottom-up strategy ensures consistent multi-level annotation.
To support structured supervision and hierarchical loss modeling, we construct two binary mapping matrices. The first, \(\mathcal{M}^1 \in \{0,1\}^{1 \times n}\), captures the mapping from a document to its sentences. The second, \(\mathcal{M}^2 \in \{0,1\}^{n \times t}\), captures the mapping from each sentence to its associated subsentences. An entry \(\mathcal{M}^2_{ij} = 1\) indicates that sentence or subsentence \(j\) belongs to sentence or document \(i\), respectively.

\subsection{ProtoSiTex Architecture}

The proposed framework, {ProtoSiTex} (Fig.~\ref{fig:arch}), consists of three core stages: (i) encoding subsentences into high-dimensional semantic representations using a language model (LM)-based encoder, (ii) employing a dual-phase learning strategy that first discovers representative prototypes through unsupervised learning, then performs supervised classification guided by hierarchical supervision using document-level taxonomy, and (iii) performing inference through prototype-based reasoning applied to the learned embeddings. We detail each component below.

\subsubsection{Subsentence Representation via LM-based Encoder} 
% \textbf{Subsentence Representation via LM-based Encoder:} 
Effective text classification relies on context-aware representations. While traditional embeddings (e.g., Word2Vec, GloVe, FastText) are static, LM-based encoders offer dynamic, context-sensitive embeddings but lack interpretability. To balance expressiveness and transparency, we use an LM-based encoder to generate subsentence embeddings.

For each \(s_{ij} \in \mathcal{S}_i\), the encoder outputs \(e_{ij} \in \mathbb{R}^d\), producing a feature matrix \(\mathcal{E} \in \mathbb{R}^{t \times d}\) over \(t\) subsentences. This matrix feeds into the prototype discovery and classification modules, enabling both semantic precision and interpretability. We next describe the dual-phase learning.

% \noindent
\subsubsection{Dual Phase Learning} 
The clustering and classification phases aim to learn prototypes and capture multi-granular document representations, respectively, as elaborated below. 

% \textbf{Clustering-phase for Prototypes Learning:}
\paragraph{\textbf{Clustering-phase for Prototypes Learning}}  
The first stage of our dual-phase learning aims to discover semantically meaningful and interpretable prototypes that facilitate multi-label classification. To achieve this, we introduce a learnable prototype layer 
% in conjunction 
coupled with a multi-head attention (MHA) mechanism, enabling diverse alignment between subsentence embeddings and prototypes.

We first initialize \( q \) prototypes \(\mathcal{P} = \{p_1, p_2, \ldots, p_q\};~ \forall p_i \in \mathbb{R}^d\).
% using random initialization, which empirically outperforms K-means, Xavier, and Kaiming. The value of \( q \) is selected via grid search.
%
Subsentence embeddings \(\mathcal{E} \in \mathbb{R}^{t \times d}\) are then projected into the prototype space using MHA, which learns flexible, context-aware alignments. For each attention head \(i\), we compute query {\scriptsize{\(\mathcal{Q}_i  = \mathcal{P}{\cdot}\mathcal{W}^{{Q}_i}\)}}, key {\scriptsize{\( \mathcal{K}_i  = \mathcal{E}{\cdot}\mathcal{W}^{{K}_i}\)}}, and value {\scriptsize{\(\mathcal{V}_i  = \mathcal{E}{\cdot}\mathcal{W}^{{V}_i} \)}} matrices,
% \begin{equation}\label{eq:head_linear_transform}
% \scriptsize
%     \begin{aligned}
%         \mathcal{Q}_i  = \mathcal{P}{\cdot}\mathcal{W}^{{Q}_i}; \quad
%         \mathcal{K}_i  = \mathcal{E}{\cdot}\mathcal{W}^{{K}_i}; \quad
%         \mathcal{V}_i  = \mathcal{E}{\cdot}\mathcal{W}^{{V}_i}
%     \end{aligned}
% \end{equation}
where {\scriptsize{\(\mathcal{W}^{Q_i}, \mathcal{W}^{K_i}, \mathcal{W}^{V_i} \in \mathbb{R}^{d \times d}\)}} are learnable parameters. Scaled dot-product attention computes the alignment:
\begin{equation}\label{eq:prototype_repr}
    \scriptsize
    % \footnotesize
    \begin{aligned}
        \mathcal{H}_i = \sigma\left( \left(\mathcal{Q}_i{\cdot}\mathcal{K}_i^\top\right)/ \sqrt{d} \right){\cdot}\mathcal{V}_i; \qquad
         \mathcal{H} = \left(\mathcal{H}_1 ~\lVert~ \mathcal{H}_2 ~\lVert ... \lVert~ \mathcal{H}_h \right){\cdot}\mathcal{W}^h
    \end{aligned}
\end{equation}
{where, $\sigma$ denotes softmax function.}
The resulting matrix {\(\mathcal{H} \in \mathbb{R}^{q \times d}\)} encodes prototype-aware representations of subsentences, which are used in the subsequent classification phase to associate prototypes with class labels.

\paragraph{\textbf{Classification-phase for Learning Multi-granular Document Details}} 

In the classification phase, we learn hierarchical label predictions using prototype embeddings $\mathcal{P} \in \mathbb{R}^{q \times d}$, subsentence embeddings $\mathcal{E} \in \mathbb{R}^{t \times d}$, and prototype-aware representations $\mathcal{H} \in \mathbb{R}^{q \times d}$. To improve generalization and interpretability, we employ multi-granular supervision without introducing separate classifiers, enabling two-way explainability: (i) subsentences $\rightarrow$ prototypes, and (ii) prototypes $\rightarrow$ class labels.

\noindent
\textbf{Prototype-to-subsentence assignment:}
We first compute the similarity between prototypes and subsentence embeddings:
\begin{equation}
\footnotesize
\mathcal{X} = \mathcal{P}{\cdot}\mathcal{E}^\top / (\lVert \mathcal{P} \rVert \lVert \mathcal{E}^\top \rVert) \in \mathbb{R}^{q \times t} .
% \mathcal{X} = \frac{\mathcal{P}\mathcal{E}^\top}{\|\mathcal{P}\|  \|\mathcal{E}^\top\|}
% \in \mathbb{R}^{q \times t}.
\end{equation}
This similarity matrix is binarized to obtain the prototype-to-subsentence
assignment matrix $\mathcal{Y} \in \{0,1\}^{q \times t}$:
\begin{equation}
\footnotesize
\mathcal{Y}_{ij} =
\begin{cases}
1, & \text{if } i = \arg\max_{k \in \{1,\dots,q\}} \mathcal{X}_{kj}, \\
0, & \text{otherwise}.
\end{cases}
\end{equation}

\noindent
\textbf{Dynamic ground-truth construction:}
Using $\mathcal{Y}$ and the subsentence label matrix $\mathcal{C} \in \{0,1\}^{t \times l}$, we construct prototype-level
ground-truth labels:
\begin{equation}
\footnotesize
\mathcal{G}_1 = \mathcal{Y}\mathcal{C} .
\end{equation}

Sentence-level ground-truth labels are obtained by aggregating subsentence labels using the sentence-to-subsentence mapping matrix $\mathcal{M}^2 \in \{0,1\}^{n \times t}$:
\begin{equation}
\footnotesize
\mathcal{G}_2 = f_{\text{clamp}}(\mathcal{M}^2 \mathcal{C}) .
\end{equation}
The function $f_{\text{clamp}}(\cdot)$ performs element-wise clipping to the $[0,1]$ interval and serves as a continuous relaxation of the logical OR (set union) operator used in multi-label aggregation. Specifically, matrix products such as $\mathcal{M}^2 \mathcal{C}$ compute the count of label occurrences across subsentences. The clamping operation maps these
counts to bounded presence indicators, where any non-zero value is saturated to $1$. Formally, this corresponds to approximating the union operation by $f_{\text{clamp}}(.)$, preserving monotonicity while ensuring numerical stability and compatibility with gradient-based optimization. This formulation enables efficient hierarchical aggregation
without introducing non-differentiable operators. 

The document-level ground truth $\mathcal{G}_3 \in \mathbb{R}^{1 \times l}$ is directly obtained from the dataset.

\noindent
\textbf{Prototype-to-label prediction:}
To map prototypes to class labels, we transform the prototype-aware representations $\mathcal{H}$ through a feed-forward network:
\begin{equation}
\footnotesize
\tilde{\mathcal{G}}_1 = \sigma\left(\mathcal{W}_3 \cdot \mathrm{ReLU}\left( \mathcal{W}_2 \cdot \mathrm{ReLU}\left(\mathcal{W}_1 \cdot \mathcal{H} \right)\right)\right), \quad \tilde{\mathcal{G}}_1 \in \mathbb{R}^{q \times l}.
\end{equation}

\noindent
\textbf{Hierarchical prediction aggregation:}
Using $\tilde{\mathcal{G}}_1$ and the assignment matrix $\mathcal{Y}$, we compute sentence-level predictions:
\begin{equation}
\footnotesize
\tilde{\mathcal{G}}_2 = f_{\text{clamp}}\left(\mathcal{M}^2 \mathcal{Y}^\top \tilde{\mathcal{G}}_1\right), \quad \tilde{\mathcal{G}}_2 \in \mathbb{R}^{n \times l}.
\end{equation}
Finally, document-level predictions are obtained via the document-to-sentence mapping matrix $\mathcal{M}^1 \in \{0,1\}^{1 \times n}$:
\begin{equation}
\footnotesize
\tilde{\mathcal{G}}_3 = f_{\text{clamp}}\left(\mathcal{M}^1 \tilde{\mathcal{G}}_2\right), \quad \tilde{\mathcal{G}}_3 \in \mathbb{R}^{1 \times l}.
\end{equation}
Both $\mathcal{G}_1$ and $\mathcal{G}_2$ are dynamically generated and depend on the current prototype configuration. This enables adaptive supervision, allowing the model to refine prototype–label associations as training evolves.

\subsubsection{Prototype Projection and Label Classification} 
Each prototype ${\mathcal{P}}_i \in \mathcal{P}$ is a high-dimensional vector residing in the same semantic space as subsentence embeddings. To ensure interpretability, we associate each prototype with its top-$K$ most similar subsentences from the training set, using cosine similarity.
The top-$K$ nearest subsentences $\mathcal{T}_K({\mathcal{P}}_i)$ are consolidated and passed to Gemini 
Pro-2.5 \cite{gemini2023}
% via an API \cite{gemini2023} 
to generate a concise summary reflecting their shared theme. 
This strategy ensures each prototype is grounded in semantically diverse and representative text snippets, offering robust explanations. 
%The prompts used for summary generation are provided in Appendix~\ref{APP:prompt}.

Given a test document $\mathcal{D}$, we segment it into subsentences, and encode them using the LM-based encoder to get $\mathcal{E} {\in} \mathbb{R}^{t \times d}$. These embeddings are projected to the prototype space via the trained multi-head attention module to produce prototype-aware representations $\mathcal{H}$, which are then passed through the proto-to-label classifier $\tilde{\mathcal{G}}_1$. Sentence- and document-level predictions ($\tilde{\mathcal{G}}_2$, $\tilde{\mathcal{G}}_3$) are obtained via hierarchical mappings using $\mathcal{M}^2$ and $\mathcal{M}^1$. The final predicted class labels for the document are taken as $\tilde{\mathcal{G}}_3$.
Simultaneously, we generate textual explanations by linking each subsentence to its closest prototype, and retrieving that prototype’s associated exemplars $\mathcal{T}_K({\mathcal{P}}_i)$. This enables transparent reasoning behind predictions, making ProtoSiTex both accurate and interpretable.

ProtoSiTex is best described as semi-interpretable because it combines prototype-based explanations with deep neural components. On the one hand, its adaptive prototypes provide human-understandable evidence by aligning subsentences with semantically coherent exemplars, allowing users to see why particular text spans support specific class predictions. On the other hand, the model relies on latent mechanisms, multi-head attention, embedding transformations, and hierarchical aggregation across subsentence, sentence, and document levels, which are not directly transparent to human reasoning. As a result, the explanations are faithful and intuitive at the prototype level but do not offer full interpretability of the entire decision process. This balance between interpretability and accuracy justifies labeling the model as semi-interpretable rather than fully interpretable.

% \textbf{\emph{Dual-phase Training Strategy of ProAspect:}}
% \subsection{Dual Phase Training Strategy}
\subsection{Training Strategy of ProtoSiTex} 
We adopt an alternate dual-phase training strategy, where each epoch comprises $i_1$ clustering and $i_2$ classification iterations. Prototype parameters are updated in the clustering phase (with classifiers frozen), and vice versa in the classification phase. This decoupling stabilizes training, prevents early overfitting, and allows prototypes to evolve independently before being used for label prediction. We now describe the loss functions associated with each phase.

% \subsubsection{Hybrid Training Strategy:}
% To train our framework, we follow a hybrid learning approach, which follows a dual-phase learning approach, clustering followed by a classification phase training. In simple terms, in a single epoch, the first $i_1$ number of iterations is run for the clustering, followed by the $i_2$ number of iterations for classification. It should be noted that during the clustering phase, the learnable parameters of the classification phase remain frozen, while only clustering-phase parameters will update, and vice versa. We now discuss the loss functions of each phase for training purposes.

\subsubsection{Loss Functions for Classification Phase}
% \textbf{\emph{Loss functions for Classification Phase}}:
The classification phase is designed to facilitate hierarchical supervision by capturing semantic patterns from fine-grained (subsentence) to coarse-grained (document) levels. To train this phase, we use three loss components: (i) a prototype-to-label interpretability loss that reinforces the semantic alignment between prototypes and classes, (ii) a sentence-wise supervision loss that captures mid-level contextual cues, and (iii) a document-level loss leveraging the provided annotations. Each loss component is computed using binary cross entropy \cite{PyTorch}, and the combined loss is expressed as:
\begin{equation}\label{eq:cs_loss}
    % \scriptsize
    \footnotesize
    \mathcal{L}_{cs} = \lambda_1 ~ \mathcal{L}_1(\mathcal{G}_1, \tilde{\mathcal{G}}_1) + 
                        \lambda_2 ~ \mathcal{L}_2 (\mathcal{G}_2, \tilde{\mathcal{G}}_2) + 
                        \lambda_3 ~ \mathcal{L}_3 (\mathcal{G}_3, \tilde{\mathcal{G}}_3)
\end{equation}

\noindent
where, $\mathcal{G}_i$ and $\tilde{\mathcal{G}}_i$ (for $i = 1, 2, 3$) represent ground-truth and predicted labels at prototype, sentence, and document levels. The weights $\lambda_i$ control each loss term, with $\sum_{i=1}^3 \lambda_i = 1$. 
Subsentence and prototype-level annotations are optional; when unavailable, 
% or unreliable, 
$\lambda_1$ and $\lambda_2$ can be reduced or set to zero, enabling the model to rely solely on document-level labels and support flexible supervision.

\subsubsection{Loss Functions for Clustering Phase} 
The clustering loss comprises two complementary components: 
\emph{proximity-based losses} that enforce alignment between subsentence embeddings and prototypes, and 
\emph{structure regularization losses} that ensure sparsity and diversity among the learned prototypes.

% \textbf{\emph{(i) Proximity-based Losses:}}  
\paragraph{\textbf{Proximity-based Losses}} 
These losses promote semantic alignment and coverage. 
The \emph{prototype matching loss} ensures that each subsentence $\mathcal{E}_i$ is closely aligned to its nearest prototype $\mathcal{P}_{j^*_i}$, encouraging relevance. Simultaneously, the \emph{interpretability loss} ensures each prototype remains active by encouraging proximity to at least one subsentence, thus achieving coverage and preventing prototype collapse. The joint formulation is given by:
% \begin{equation}\label{eq:proximity_loss}
%     % \scriptsize 
%     \footnotesize
%     \begin{aligned}
%     \mathcal{L}_{prox} =  
%     -\frac{1}{t} \sum_{i=1}^{t} \left(\frac{\mathcal{E}_i^\top \mathcal{P}_{j^*_i}}{\lVert\mathcal{E}_i\rVert \lVert\mathcal{P}_{j^*_i}\rVert} \right)
%      +  \frac{1}{q} \sum_{j=1}^{q} \left(1 -  \frac{\mathcal{E}_i^\top \mathcal{P}_{j^*_i}}{\lVert\mathcal{E}_i\rVert \lVert\mathcal{P}_{j^*_i}\rVert} \right)
%     \end{aligned}          
% \end{equation}
\begin{equation}\label{eq:proximity_loss}
    % \scriptsize 
    \footnotesize
    \begin{aligned}
    \mathcal{L}_{prox} =  
    -\frac{1}{b} \sum_{i=1}^{b} \left(\frac{\mathcal{E}_i^\top \mathcal{P}_{j^*_i}}{\lVert\mathcal{E}_i\rVert \lVert\mathcal{P}_{j^*_i}\rVert} \right)
     +  \frac{1}{q} \sum_{j=1}^{q} \left(1 -  \max_{i = 1}^b \left(\frac{\mathcal{E}_i^\top \mathcal{P}_{j}}{\lVert\mathcal{E}_i\rVert \lVert\mathcal{P}_{j}\rVert}\right) \right)
    \end{aligned}          
\end{equation}
where $j^*_i$ is the index of the prototype closest to subsentence $\mathcal{E}_i$ based on cosine similarity, and $b$ denotes the total number of subsentences across all documents in a mini-batch. 

% \textbf{\emph{(ii) Structure Regularization Losses:}}  
\paragraph{\textbf{Structure Regularization Losses}} 
To enhance interpretability, we impose \emph{sparsity} and \emph{diversity} through regularization. The sparsity loss encourages selective activation, ensuring a limited set of subsentences influences each prototype. This reduces overlap and enhances clarity. The diversity loss, implemented via orthogonality regularization, pushes prototypes apart in embedding space, minimizing redundancy. The combined regularization is:
\begin{equation}\label{eq:reg_loss}
    % \scriptsize
    \footnotesize
    \begin{aligned}
    \mathcal{L}_{reg} = -\frac{1}{q} \sum_{j=1}^{q} \log\left( \sum_{i=1}^{b} \frac{\exp(-\left\lVert \mathcal{E}_i - \mathcal{P}_j \right\rVert_2)}{\sum_{k=1}^{q} \exp(-\left\lVert \mathcal{E}_i - \mathcal{P}_k \right\rVert_2)} + \epsilon \right) \\
    + 
    \frac{1}{q(q-1)} \sum_{\substack{i,j=1; \\ i \ne j}}^{q} \left( \frac{\mathcal{P}_i^\top \mathcal{P}_j}{\lVert\mathcal{P}_i\rVert \lVert\mathcal{P}_j\rVert} \right)^2
    \end{aligned}          
\end{equation}
where, the first term encourages peaked assignment distributions (sparsity), and the second term enforces pairwise dissimilarity (diversity) between prototypes.

% \noindent
To enforce semantic alignment with classes, we augment the clustering loss with the supervised classification loss $\mathcal{L}_{cs}$. The final clustering-phase loss ($\mathcal{L}_{cr}$) is:
\begin{equation}\label{eq:cl_loss}
    % \scriptsize
    \footnotesize
    \mathcal{L}_{cr} = \alpha_1 ~ \mathcal{L}_{prox} + \alpha_2 ~\mathcal{L}_{reg} + \alpha_3 ~ \mathcal{L}_{cs}
\end{equation}
where, $\alpha_1, \alpha_2, \alpha_3$ are tunable weights summing to 1, controlling the influence of each component.

The classification phase refines prototypes into interpretable classes, 
%complementing 
enhancing clustering and enabling generalizable, multi-granular predictions. 
% The next section outlines the prototype-based inference process.
% Next, we describe the prototype-based inference process.

The computational complexity of {ProtoSiTex} is analyzed in Appendix \textcolor{blue}{A} of the supplementary file. Appendix \textcolor{blue}{B} provides a detailed discussion on the stability of prototype updates and the influence of loss-weight parameters on model behavior.

\section{Experiments}\label{sec:experiments}

% This section describes the datasets employed and reports the corresponding experimental results.

\subsection{Datasets} 
To evaluate the effectiveness of ProtoSiTex and compare it against state-of-the-art (SOTA) methods, we conducted comprehensive experiments on benchmark datasets IMDb \cite{IMDb}, TweetEVAL \cite{TweetEVAL}, and our newly proposed hotel reviews dataset (HR). These datasets encompass a diverse range of classification tasks: binary, multi-class, and multi-label, respectively.

\emph{\textbf{(i) IMDb}} \cite{IMDb}: 
Consisting of 50K movie reviews for binary sentiment classification (positive/ negative).

\emph{\textbf{(ii) TweetEVAL}} \cite{TweetEVAL}:
A multiclass dataset comprising 5052 tweets, with each tweet annotated with one of four emotion labels: anger, joy, optimism, or sadness.

\definecolor{price}{HTML}{4E6FDA}
\definecolor{person}{HTML}{ED4C3B}
\definecolor{concierge}{HTML}{F0CD38}
\definecolor{food}{HTML}{45C35D}
\definecolor{room}{HTML}{EC7D0E}
\definecolor{house}{HTML}{4CC7F2}
\definecolor{location}{HTML}{8F85CD}
\definecolor{guest}{HTML}{CA88A6}

% \begin{table}[!t]
% \centering
% % \scriptsize
%     % \centering
    
%     \caption{Proposed Dataset: Hotel Review (HR)}
%     \label{tab:dataset_overview}
% \begin{adjustbox}{width=0.35\textwidth}
%     \begin{tabular}{ll}
%    \hline
%    \textbf{Attributes} & \textbf{Values} \\ \hline \hline 
%    No. of Hotels \& Locations  & 275 \& 70  \\
%    No. of Reviews \& Sentences  & 3002 \& 26519  \\
%    No. of Aspects & 8 \\
%    (Min. \& Max.) sentence per review   & 1 \& 100   \\ 
%    (Avg. \& Std. Dev.) sentence per review  & 8.83 \& 7.47   \\ 
%    Average review length    & 102.39   \\\hline
%     \end{tabular}
% \end{adjustbox}
% \end{table}

%%%%%%%%%%%%%%%%%%%%%%%%%%%%%%%%%%%%%%%%%

%%%%%%%%%%%%%%%%%%%%%%%%%%%%%%%%%%%%%%%%%
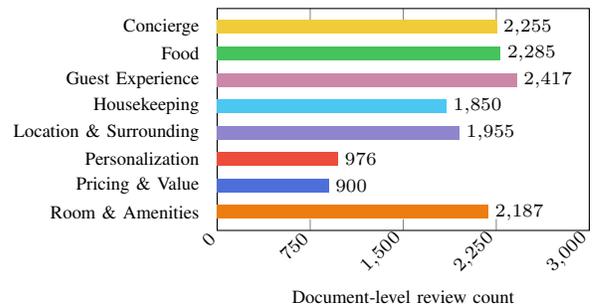
\begin{figure}[!b]
    \centering
    \begin{subfigure}{0.48\textwidth}
    % \tiny
    \scriptsize
       \centering
       \begin{tikzpicture}
      \begin{axis}[
         width=0.75\textwidth,
         % height=2\textwidth,
         xbar,
         bar width=5pt,
         xmin=0,
         xmax=3000,
         xtick={0,750,1500,2250,3000},
         ytick={
           Room \& Amenities,
           Pricing \& Value,
           Personalization,
           Location \& Surrounding,
           Housekeeping,
           Guest Experience,
           Food,
           Concierge,
         },
         symbolic y coords={
           Room \& Amenities,
           Pricing \& Value,
           Personalization,
           Location \& Surrounding,
           Housekeeping,
           Guest Experience,
           Food,
           Concierge,
         },
         y=10pt, 
         yticklabel style={
           % font=\tiny,     % larger font for better readability
           align=right,     % align labels right
         },
         % nodes near coords,
         nodes near coords align={horizontal},
         nodes near coords,
         nodes near coords style={text=black, anchor=west},
         xticklabel style={rotate=45, anchor=east},   
         % height=0.45\textwidth,
         xlabel={Document-level review count},
         enlarge y limits,
         ytick style={draw=none,}
          ]
          \addplot+[xbar, fill=room, draw opacity=0, bar shift=0pt]   coordinates {(2187,Room \& Amenities)};
          \addplot+[xbar, fill=price, draw opacity=0, bar shift=0pt] coordinates {(900,Pricing \& Value)};
          \addplot+[xbar, fill=person, draw opacity=0, bar shift=0pt] coordinates {(976,Personalization)};
          \addplot+[xbar, fill=location, draw opacity=0, bar shift=0pt]   coordinates {(1955,Location \& Surrounding)};
          \addplot+[xbar, fill=house, draw opacity=0, bar shift=0pt]    coordinates {(1850,Housekeeping)};
          \addplot+[xbar, fill=guest, draw opacity=0, bar shift=0pt]     coordinates {(2417,Guest Experience)};
          \addplot+[xbar, fill=food, draw opacity=0, bar shift=0pt]   coordinates {(2285,Food)};
          \addplot+[xbar, fill=concierge, draw opacity=0, bar shift=0pt] coordinates {(2255,Concierge)};
           \end{axis}
            \end{tikzpicture}
    \end{subfigure}
\caption{Distribution of eight aspects in HR}
\label{fig:hr_dataset_distribution}
\end{figure}

\emph{\textbf{(iii) Proposed Dataset on Hotel Reviews (HR)}}:  
To support fine-grained, interpretable, multi-aspect classification in the hospitality domain, we introduce a novel dataset comprising hotel reviews, curated from \emph{tripadvisor.com}. The reviews were collected via web scraping, ensuring a diverse coverage across locations, 
% \textcolor{red}{hotel categories}, 
and customer sentiments. 
We identified $8$ key aspects based on extensive consultations with experts in travel management and linguistics (Fig. \ref{fig:hr_dataset_distribution}). These aspects were chosen to reflect both operational and experiential dimensions of hotel service quality. 
Each review may correspond to multiple aspect labels, reflecting the inherently multi-label nature of user opinions. To ensure high-quality annotations, we employed $10$ trained human annotators for each review. For every review, the annotators assigned one or more aspect labels as applicable, and highlighted the exact subsentence or phrase responsible for each aspect label, thereby providing ground-truth supervision for interpretability evaluation.
Prior to annotation, all annotators underwent training. 
% To ensure annotation consistency and reliability, 
We implemented a quality control mechanism using in-context learning via GPT-4o \cite{ARTICLE_AAAI_2025}. Annotator performance was iteratively assessed by comparing human-labeled segments, enabling correction and refinement through active feedback loops. 
To ensure annotation consistency and reliability, we quantified inter-annotator agreement using Krippendorff’s alpha ($\upalpha$) \cite{krippendorff1970}. 
% The obtained $\upalpha=0.85$ signifies strong agreement among annotators, and supports the annotation procedure. 
The obtained $\upalpha = 0.85$ indicates strong inter-annotator agreement, validating the annotation procedure. 
The resulting dataset consists of 3002 document-level and 26519 sentence-level reviews, annotated at the subsentence level with multi-label aspect tags and text spans responsible for each label, making it one of the first of its kind to support interpretability-grounded multi-aspect classification at such fine granularity.
Brief statistical details of HR are presented in Table \ref{tab:dataset_overview}. 

Notably, subsentence level annotations are absent in the IMDb and TweetEVAL datasets.

\subsection{Experimental Setup}
All experiments were conducted on a Dell Precision 5860 Tower equipped with PyTorch 2.5.1, CUDA 12.2, an NVIDIA A100 GPU with 80 GB VRAM, and 196 GB RAM. 
% \textcolor{red}{We employed RoBERTa-Large \cite{liu2019roberta} as the backbone for encoding subsentences in our framework.} 
% The model was trained using AdamW \cite{ADAMW} with a learning rate of $2{\times}10^{-5}$. 
The HR dataset was split into training, validation, and test sets in $7{:}1{:}2$. For IMDb \cite{IMDb} and TweetEVAL \cite{TweetEVAL}, we used the given training/ validation/ test sets. Each experiment was run five times, and the mean test performance is reported here.
% Additional model configuration details are provided in the  Appendix {A}. 
%\ref{APP:Hyperparameters}.
%\emph{supp. file}.
% All reported results correspond to the test set. 

Table~\ref{tab:hyperparameter} presents the hyperparameters engaged in ProtoSiTex. To obtain high-dimensional, contextually rich text embeddings, we incorporate a language model, called RoBERTa \cite{liu2019roberta}. Specifically, it is a robustly optimized BERT variant, which we employed in its large-weight version pretrained on large English-language internet corpora using the masked language modeling objective; RoBERTa-Large remains tunable within our framework. We maintain a consistent dropout rate of 0.1 across all trainable weights. To avoid overfitting, early stopping with a patience of three epochs based on validation performance is applied. Additionally, we set a maximum training limit of 100 epochs. Each epoch includes 10 ($=i_1$) iterations of the clustering phase and 100 ($=i_2$) iterations of the classification phase, ensuring a balanced optimization of both phases.

\begin{table}[!t]
\centering
\caption{Statistics of proposed dataset: Hotel Reviews (HR)}
\label{tab:dataset_overview}
\begin{adjustbox}{width=0.37\textwidth}
    \begin{tabular}{l|c}
        \hline 
        \textbf{Attribute} & \textbf{Value} \\ \hline 
        No. of hotels   \& locations  & 275 \& 70 \\
        No. of reviews   \& sentences & 3002 \& 27074  \\
        No. of aspects & 8 \\
        Sentences per review   (min \& max) & 1 \& 92   \\
        Sentences per review   (mean $\pm$ stdev) & 9.01 $\pm$ 7.69    \\
        % Word per review (Avg.) & 120.82  \\
        % Words per review (mean $\pm$ stdev) & 120.82 $\pm$ \textcolor{red}{122.62}  \\
        Words per review (mean) & 120.82  \\
        \hline 
    \end{tabular}
\end{adjustbox}
\end{table}

\begin{table}[!b]
    \centering
    \small
    \caption{Brief details of hyperparameters}
    \begin{adjustbox}{width=0.43\textwidth}
    \begin{tabular}{l|c}
        \hline
        \textbf{Attribute}    & \textbf{Value}  \\ \hline
        Optimizer     & AdamW                               \\ 
         Learning rate (Clustering-phase) & $2 \times 10^{-5}$ \\
         Learning rate (Classification-phase) & $2 \times 10^{-5}$  \\
        Batch size    & 64                             \\ \hline
        LM encoder: Feature dimension & 1024 \\ \hline
        No. of dense layer  & 2 \\
        Dimension of dense layer & 256 \& 64 \\ \hline
        Clustering loss parameters ($\alpha_1$, $\alpha_2$, $\alpha_3$ )        & [0-1] \\ 
        Classification loss parameters ($\lambda_1$, $\lambda_2$, $\lambda_3$) &       [0-1]              \\  \hline
        No. of attention heads ($h$) & [1-32] \\ \hline
        No. of learnable prototypes ($q$) & [4-64] \\ \hline
    \end{tabular}
    \label{tab:hyperparameter}
    \end{adjustbox}
\end{table}

\begin{table*}[!hbt]
    \centering
    % \scriptsize
    % \caption{Comparison with baseline across nine black-box methods, and five SOTA approaches}
% \caption{Comparison with baseline and SOTA approaches}
\caption{Comparison with baseline approaches}
\begin{adjustbox}{width=0.8\textwidth}
\begin{tabular}{c|c|l|c|c|c|c|c|c|c}
\cline{4-10}
\multicolumn{1}{c}{} & \multicolumn{1}{c}{} & \multicolumn{1}{c}{} & \multicolumn{3}{c|}{\textbf{Hotel Reviews (HR)} } & \multicolumn{2}{c|}{\textbf{IMDb} \cite{IMDb} } & \multicolumn{2}{c}{\textbf{TweetEVAL} \cite{TweetEVAL}}  \\ \cline{2-10} 

\multicolumn{1}{c}{} & \multicolumn{1}{c|}{\textbf{Category}}  &  \multicolumn{1}{l|}{\textbf{Model}} & {\textbf{${\mathcal{F}}_{m}\uparrow$}} & {\textbf{${\mathcal{BA}}_{m}\uparrow$}} & {\textbf{${\mathcal{HL}}\downarrow$}}  & {\textbf{${\mathcal{A}}\uparrow$}} & {\textbf{${\mathcal{F}}\uparrow$}}  & {\textbf{${\mathcal{A}}\uparrow$}} &  {\textbf{${\mathcal{F}}\uparrow$}}  \\ \hline

% \multirow{3}{*}{\rotatebox{90}{TC}}
% % & M1 &  &  &  &  &  &  &  \\

\multirow{30}{*}{\rotatebox{90}{Black-box Models}}

%======= Traditional
& \multirow{3}{*}{\shortstack{Traditional Text\\Classification}}
& Word2Vec \cite{word2vec} + BiLSTM & 61.72 & 58.15 & 0.2647 & 86.20 & 86.18 & 60.87 & 55.79 \\
& & FastText \cite{fasttext} + BiLSTM & 61.32 & 58.90 & 0.2720 & 81.93 & 81.39 & 59.81 & 56.80 \\
& & Glove \cite{glove} + BiLSTM & \underline{64.74} & \underline{62.36} & \underline{0.2304} & \underline{86.95} & \underline{86.93} & \underline{63.35} & \underline{59.75} \\
\cline{2-10}

%======= Text Classification Methods

&\multirow{2}{*}{\shortstack{Graph-based Text\\Classification}}
& TextGCN \cite{textgcn} & 78.37 & 64.07 & 0.2288 & 87.29 & 87.27 & 62.14 &  61.36\\
& & BertGCN \cite{bertgcn} & \underline{86.55} & \underline{74.56} & \underline{0.1612} & \underline{93.72} & \underline{93.71} & \underline{80.43} & \underline{80.80} \\
\cline{2-10}

&\multirow{7}{*}{\shortstack{Multi-label Text\\Classification}} 
&  Binary Relevance \cite{binary_relevance} & 70.81 & \underline{66.94} & 0.3159 & 74.74 & 75.52 & 59.95 & 59.11\\
& & Classifier Chain \cite{classifier_chain} & 70.66 & 66.83 & 0.3178 & 74.74 & 75.52 & 55.59 & 52.86 \\
& & Label power set \cite{label_power_set} & 72.79 & 62.90 & 0.2805 & 74.74 & 75.52 & 60.89 & 60.02 \\
& & ML-KNN \cite{ml_knn}& 72.09 & 60.65 & 0.2814 & 73.61 & 76.11 & 55.31 & 54.95 \\
& & MAGNET \cite{MAGNET}& \underline{76.42} & 56.01 & 0.2747 & 86.26 & 86.26 & 72.27 & 70.39 \\
& & UCLAF \cite{UCLAF} & 73.30 & 61.99 & \underline{0.2341} & \underline{90.82} & \underline{90.97} & \underline{76.28} & \underline{76.04} \\
& & MLTC$_1$ \cite{tkde_26} & 74.43 & 51.15 & 0.3662 & 82.59 & 82.72 & 33.07 & 33.57\\
\cline{2-10}

&\multirow{4}{*}{\shortstack{SOTA for\\ TweetEVAL and \\IMDB}} 
& BERTweet \cite{bertweet} & 80.10 & \underline{76.94} & 0.1801 & 89.60 & 89.95 & 83.95 & 83.80 \\
& & TweetNLP \cite{tweetnlp} & \underline{80.44} & 76.28 & \underline{0.1789} & 90.46 & 90.71 & \underline{84.09} & \underline{83.90} \\
& & SETFIT \cite{SETFIT} & 72.23 & 70.55 & 0.1801 & 91.23 & 91.35 & 70.09 & 71.07 \\
& & SentiCSE \cite{SentiCSE} & 69.22 & 66.92 & 0.2246 & \underline{94.03} & \underline{94.01} & 76.07 & 75.68 \\
\cline{2-10}

%======== Masked-based Models
% \hline
& \multirow{11}{*}{\shortstack{Masked-based\\Models}}
& ALBERT \cite{ALBERT} & 83.73 & 77.54 & 0.1477 & 90.29 & 90.12 & 75.58 & 74.98 \\
& & BERT \cite{devlin-etal-2019-bert} & 86.16 & 82.01 & 0.1253 & 92.62 & 92.49 & 79.73 & 79.39 \\ 
& & DeBERTaV3 \cite{DeBERTaV3} & 86.24 & 80.09 & 0.1385 & 95.40 & 95.39 & 81.63 & 81.38 \\
& & DistilBERT \cite{sanh2019distilbert} & 84.96 & 81.14 & 0.1411 & 92.25 & 92.16 & 79.59 & 79.16 \\ 
& & ELECTRA \cite{ELECTRA} & 86.78 & 81.16 & 0.1356 & 95.07 & 95.11 & 81.14 & 80.92 \\
& & ModernBERT \cite{ModernBERT} & 85.60 & 83.21 & 0.1411 & 95.39 & 95.36 & 73.18 & 73.16 \\ 
& & ModernBERT-Large \cite{ModernBERT} & 86.76 & 82.36 & 0.1337 & 96.08 & 96.11 & 77.76 & 77.33 \\
& & Nomic Embed Text V2 \cite{nomic_embed} & 83.06 & 77.41 & 0.1506 & 93.67 & 93.67 & 81.91 & 81.49 \\
& & NeoBERT \cite{neobert} & 86.26 & 82.79 & 0.1292 & 94.70 & 94.82 & 80.86 & 80.59\\ 

& & RoBERTa \cite{liu2019roberta} & 86.47 & 83.41 & 0.1277 & 93.74 & 93.82 & 80.78 & 80.66 \\ 
& & RoBERTa-Large \cite{liu2019roberta} & \underline{87.23} & \underline{84.59} & \underline{0.1204} & \underline{96.34} & \underline{96.32} & \underline{83.18} & \underline{82.83} \\

\cline{2-10}

%========= Autoregressive Models
& \multirow{4}{*}{\shortstack{Autoregressive\\Models}}
& XLNet \cite{yang2019xlnet} & \underline{86.95} & \underline{83.51} & \underline{0.1325} & 94.68 & 94.79 & 77.20 & 77.11 \\ 
& & Qwen3-4B \cite{qwen3}& 81.34 & 76.54 & 0.1807 & 96.33 & 96.33 & 78.68 & 78.48 \\
& & Llama3.2-3B \cite{llama3} & 84.14 & 80.37 & 0.1502 & \underline{96.72} & \underline{96.75} & \underline{82.69} & \underline{82.38}\\
& & Phi-4-mini-instruct \cite{phi_4} & 80.31 & 75.20 & 0.1870 & 96.06 & 96.07 & 79.24 & 79.05\\
\hline
% \multicolumn{2}{c|}{Improvement (\%)} & -0.72 & -1.20& -0.33 & 0.32 & 0.32 & 1.36 & 1.59 \\ \hline
% \hline

% \multicolumn{2}{c|}{Improvement (\%)} & -0.92 & -2.11& -0.35 & 0.24 & 0.26 & 1.19 & 1.56 \\ \hline
% \hline

\multirow{12}{*}{\rotatebox{90}{Explainable Models}}
%================= Closed Source LLMs
& \multirow{3}{*}{\shortstack{Closed-source\\LLMs}}
& Gemini-2.5-flash \cite{gemini_2.5} & \underline{78.35} & \underline{74.95} & \underline{0.2056} & 95.79 & 95.79 & 80.08 & 80.26 \\
& & GPT-5-nano \cite{gpt_5} & 76.33 & 73.49 & 0.2150 & \underline{95.95} & \underline{95.95} & \underline{80.92} & \underline{81.44}\\
& & Deepseek-chat-v3.2 \cite{deepseek} & 72.11 & 73.37 & 0.2396 & 95.86 & 95.86 & 79.94 & 80.30 \\
\cline{2-10}

%================= Open Source LLMs
& \multirow{3}{*}{\shortstack{Open-source\\LLMs}}
& Qwen2.5-72B-instruct \cite{qwen2.5} & 61.44 & \underline{66.76} & \underline{0.3178} & 87.36 & 87.36 & 80.15 & 80.55 \\
& & Llama3.3-70B-instruct \cite{llama3} & \underline{65.16} & 63.38 & 0.3319 & \underline{88.49} & \underline{88.49} & \underline{81.98} & \underline{82.16}  \\
& & Mistral-small-3.2-24B \cite{mistralsmall3} & 61.03 & 65.77 & 0.3248 & 85.51 & 85.51& 80.64 & 79.76 \\
\cline{2-10}

%% ========= SOTA

& \multirow{6}{*}{\shortstack{Prototypical\\Models}}
& Proto-lm \cite{xie-etal-2023-proto} & 62.44 & 49.88 & 0.2894 & 90.79 & 90.78 & 22.15 & 39.26 \\ 
& & ProtoryNet \cite{10.5555/3648699.3648963-ProtoryNet} & 62.57 & \underline{50.00} & 0.2877 & 88.21 & 88.45 & 72.90 & 72.28 \\ 
& & GAProtoNet \cite{wen-etal-2025-gaprotonet} & 72.31 & 48.13 & \underline{0.2360} & \underline{96.53} & \underline{96.53} & \underline{81.84} & \underline{81.85} \\ 
& & ProSeNet \cite{10.1145/3292500.3330908-ProsNet} & 74.31 & \underline{50.00} & 0.3842 & 86.69 & 87.53 & 73.24 & 73.54 \\
& & ProtoTEX \cite{das-etal-2022-prototex} & \underline{77.24} & 38.10 & 0.2460 & 92.39 & 91.36 & 75.28 & 75.20 \\

& & ProtoLens\cite{protolens} & 74.31 & \underline{50.00} & 0.3842 & 91.10 & 91.09 & 50.59 & 51.05 \\ \hline 

% \multicolumn{2}{c|}{Improvement (\%)} & 13.16 & 68.91 & 14.73 & 0.11 & 0.10 & 3.02 & 2.81 \\ \hline
% \hline

% \multicolumn{2}{c|}{Improvement (\%)} & 12.92 & 67.36 & 14.73 & 0.04 & 0.04 & 2.85 & 2.77 \\ \hline
% \hline

%==== Ours
% \multicolumn{1}{c}{} & \textbf{ProAspect} & \textbf{87.40} & \textbf{84.46} & \textbf{87.66} & \textbf{96.64} & \textbf{96.63} & \textbf{84.31} & \textbf{84.15} \\ \hline

\multicolumn{1}{c}{} & \multicolumn{1}{c}{} & \textbf{ProtoSiTex} & \textbf{87.22$\pm$0.15} & \textbf{83.68$\pm$0.70} & \textbf{0.1235$\pm$0.06} & \textbf{96.57$\pm$0.11} & \textbf{96.57$\pm$0.10} & \textbf{84.17$\pm$0.23} & \textbf{84.12$\pm$0.03} \\ \hline
\multicolumn{10}{r}{Best method in each category are \underline{underlined}}
\end{tabular}
\end{adjustbox}
\label{tab:sota}
\end{table*}

\begin{table*}[hbt]
\centering
\caption{Performance of ProtoSiTex with different encoders}
\label{tab:different_encoder}
\begin{adjustbox}{width=0.8\textwidth}
\scriptsize
\begin{tabular}{c|l|c|c|c|c|c|c|c}
\cline{3-9}
 \multicolumn{1}{c}{} & \multicolumn{1}{c}{} & \multicolumn{3}{c|}{\textbf{Hotel Reviews (HR)} } & \multicolumn{2}{c|}{\textbf{IMDb} \cite{IMDb} } & \multicolumn{2}{c}{\textbf{TweetEVAL} \cite{TweetEVAL}}  \\ \hline

 \multicolumn{1}{c|}{\textbf{Category}}  &  \multicolumn{1}{l|}{\textbf{Model}} & {\textbf{${\mathcal{F}}_{m}\uparrow$}} & {\textbf{${\mathcal{BA}}_{m}\uparrow$}} & {\textbf{${\mathcal{HL}}\downarrow$}}  & {\textbf{${\mathcal{A}}\uparrow$}} & {\textbf{${\mathcal{F}}\uparrow$}}  & {\textbf{${\mathcal{A}}\uparrow$}} &  {\textbf{${\mathcal{F}}\uparrow$}}  \\ \hline

\multirow{3}{*}{Masked Based} & 
Nomic Embed Text V2 \cite{nomic_embed} & 81.33 & 79.41 & 0.1846 & 93.70 & 93.78 & 83.11 & 82.98 \\ 
% \cline{2-9} 
& ELECTRA \cite{ELECTRA} & 84.46 & 80.93 & 0.1360 & 95.10 & 95.11 & 81.28 & 81.24 \\ 
% \cline{2-9} 
 & RoBERTa-Large \cite{liu2019roberta} & 87.22 & 83.68 & 0.1235 & 96.57 & 96.57 & 84.17 & 84.12 \\ \hline
\multirow{3}{*}{Autoregressive} & XLNet \cite{yang2019xlnet} & 83.87 & 77.26 & 0.1593 & 94.90 & 94.91 & 77.97 & 77.82\\

& Qwen3-4B \cite{qwen3} & 76.81 & 77.37 & 0.2263 & 96.47 & 96.47 & 78.96 & 78.75 \\ 
% \cline{2-9} 
 & Llama3.2-3B \cite{llama3} & 78.52 & 79.01 & 0.2123 & 96.89 & 96.91 & 83.81 & 83.53 \\ \hline
 % \cline{2-9} 
 % & Phi-4-mini-instruct \cite{phi_4} & \textcolor{red}{58.43} & \textcolor{red}{67.42} & \textcolor{red}{0.2849} & 96.14 & 96.13 & 80.72 & 80.56 \\ \hline
\end{tabular}
\end{adjustbox}
\end{table*}

\subsection{Evaluation Metrics}

We evaluate model performance using task-specific metrics. 
For binary and multi-class classification, we report Accuracy ($\mathcal{A}$) and F1-score ($\mathcal{F}$); 
aspect-wise evaluation additionally includes Precision ($\mathcal{P}r$) and Recall ($\mathcal{R}$); 
for multi-label classification, we use macro-averaged F1-score ($\mathcal{F}_m$) and Balanced Accuracy ($\mathcal{BA}_m$); 
all are in percentages. 
We also compute Hamming loss ($\mathcal{HL}$) to analyze label-wise prediction performance, 
and report complementary Hamming loss ${\mathcal{HL}}^c$ here; 
${\mathcal{HL}}^c = (1 - \mathcal{HL}) \times 100\%$.
% We also compute Hamming Loss ($\mathcal{HL}$) and ${\mathcal{HL}}^c$ to quantify label-wise prediction accuracy; 
% ${\mathcal{HL}}^c = (1 - \mathcal{HL}) \times 100\%$.

%%%%%%%%%%%%%%%%%%%%%%%%%%%%%%%%%%%%%%%%%%%%%%%%%%%%%%%%%%%%%%%%%%%%%%%%%Result analysis
\subsection{Results \& Comparative Study}
\noindent
To comprehensively evaluate ProtoSiTex, we compare it against a diverse set of baselines and SOTA models, as summarized in Table \ref{tab:sota}.

\subsubsection{Comparison with Black-box Models}
As observed from Table \ref{tab:sota}, compared with traditional, graph-based, and advanced black-box models, {ProtoSiTex} demonstrates consistent performance improvements, highlighting the limitations of static embeddings, recurrent encoders, and global representation learning in capturing contextual semantics and fine-grained label interactions. While graph-based methods improved over traditional baselines through corpus-level structural modeling, they remained coarse-grained and relied on global propagation, limiting sensitivity to localized semantics. Similarly, classical multi-label approaches (e.g., Binary Relevance, Classifier Chains, ML-KNN) underperformed due to their reliance on label independence or shallow correlations, failing to model overlapping and context-dependent semantics. Recent state-of-the-art models such as BERTweet, TweetNLP, SetFit, SentiCSE, and transformer-based architectures (e.g., BERT, RoBERTa, DeBERTaV3, GPT, LLaMA) achieved strong performance via contextual or contrastive representation learning; however, they primarily operate on global embeddings and lack mechanisms for fine-grained, faithful explanation. In contrast, {ProtoSiTex} integrates contextual subsentence representations with adaptive prototype learning and hierarchical supervision, enabling explicit alignment between localized textual evidence and labels. This results in improved balanced accuracy and reduced Hamming loss, indicating better handling of class imbalance and more reliable label-wise predictions, while providing intrinsic, prototype-driven interpretability and a favorable trade-off between predictive performance and transparency.

\subsubsection{Comparison with Explainable Models}
Compared with closed-source (e.g., GPT-family, Gemini, DeepSeek) and open-source LLMs (e.g., LLaMA, Mistral, Qwen), {ProtoSiTex} achieved comparable performance on IMDb while consistently outperforming them on more challenging datasets such as HR and TweetEval, reflecting its strength in modeling fine-grained and multi-label semantics. Although LLMs benefit from large-scale pretraining, they are primarily optimized for generative objectives and rely on distributed representations, resulting in limited label-wise consistency and a lack of inherent mechanisms for faithful attribution. Existing prototype-based approaches provide interpretability but rely on static or weakly adaptive prototypes at coarse granularity, leading to poor alignment with labels, reduced discriminative power, and instability across datasets. In contrast, {ProtoSiTex} integrates adaptive prototype learning within a dual-phase framework and hierarchical supervision across subsentence, sentence, and document levels, enabling precise alignment between localized evidence and labels. This results in improved handling of overlapping semantics, more reliable label-wise predictions, and intrinsically interpretable, prototype-grounded explanations, achieving a principled balance between performance and transparency.

\subsubsection{Performance of ProtoSiTex}
Across datasets, ProtoSiTex demonstrates stable performance, remaining comparable on simpler tasks such as IMDb while achieving clear gains on more challenging settings like HR and TweetEval. The improvements in balanced accuracy and Hamming loss indicate better handling of class imbalance and more reliable label-wise predictions. This is driven by its dual-phase optimization and hierarchical supervision, which promote well-structured representation geometry and effective multi-label disentanglement through alignment of subsentence-level evidence with labels, resulting in improved generalization.

\subsection{Impact of Different Encoders}
Table \ref{tab:different_encoder} presents the performance of ProtoSiTex with different encoders. We consider three masked-based encoders: Nomic Embed Text V2 \cite{nomic_embed}, ELECTRA \cite{ELECTRA}, and RoBERTa-Large \cite{liu2019roberta}, and three autoregressive encoders: XLNet \cite{yang2019xlnet}, Qwen3-4B \cite{qwen3}, and LLaMA3.2-3B \cite{llama3}, as this categorization aligns with the internal working of ProtoSiTex.

ProtoSiTex depends on the quality of text representations for clustering and similarity-based prototype matching. Masked-based encoders, trained with bidirectional context, generally produce dense and semantically well-structured embeddings, making them highly suitable for ProtoSiTex. This is evident in the HR dataset, where masked-based models consistently outperformed autoregressive models. RoBERTa-Large achieved the performance (${\mathcal{F}}_{m}$=87.22), followed by ELECTRA and Nomic. In contrast, autoregressive models with unidirectional context, such as Qwen3-4B and LLaMA3.2-3B, showed notable degradation. 
However, XLNet, an autoregressive model with bidirectional context, also exhibited competitive performance. This suggests that the effectiveness of ProtoSiTex is not solely determined by the encoder category, but by the quality of text representations.

% This indicates that masked encoders better support fine-grained prototype alignment, especially in multi-label settings requiring precise semantic separation

On simpler tasks such as IMDb, the performance gap between encoder types was reduced. Autoregressive models performed competitively, with LLaMA3.2-3B (${\mathcal{A}}$=96.89) slightly outperforming RoBERTa-Large (${\mathcal{A}}$=96.57). This suggests that coarse-grained tasks depend more on overall semantic understanding. TweetEVAL shows an intermediate trend. RoBERTa-Large outperforms others, while autoregressive models remain competitive.

Overall, the results demonstrate that ProtoSiTex is more compatible with masked-based encoders, particularly for fine-grained, multi-label tasks requiring interpretability. Their structured embedding space supports effective prototype clustering and hierarchical learning. However, despite strong general language understanding, autoregressive models are less suitable for fine-grained semantic alignment. Appendix \textcolor{blue}{G} provides a detailed latent space analysis of these encoders.

% To better understand the semantic structure of learned prototypes, we visualize the prototype space using t-SNE for IMDb, TweetEVAL, and HR datasets (Fig.~\ref{fig:tsne}), where ProAspect effectively clusters semantically similar prototypes, reflecting coherent alignment with multi-labels. 

\input{plots/phases_variation}
\begin{figure}[!t]
    \centering
    \begin{subfigure}{0.23\textwidth}\tiny
        \centering
         \begin{tikzpicture}
              \begin{axis}[
                width=\textwidth,
                height=\textwidth,
                view={135}{35},
                xlabel={$\alpha_1$},
                ylabel={$\alpha_2$},
                zlabel={$\alpha_3$},
                xtick={0, 0.25, 0.50, 0.75, 1.0},
                ytick={0, 0.25, 0.50, 0.75, 1.0},
                ztick={0, 0.25, 0.50, 0.75, 1.0},
                xticklabel style={rotate=90, anchor=east},
                yticklabel style={rotate=45, anchor=east},
                % colormap/viridis,
                colormap/jet,
                colorbar,
                point meta min=67,
                point meta max=88,
                enlarge x limits={0.1},
                enlarge y limits={0.1},
                enlarge z limits={0.1},
                colorbar style={
                    width=0.15cm,
                    ytick={67,70,73,76,79,82,85,88},
                    yticklabel style={font=\tiny},
                    title= $\mathcal{F}_m$,
                    title style={yshift=-2ex, font=\tiny},
                  },
                only marks,
                mark=*,
                mark size=1.5pt,
                draw=black,
                line width=0.1pt,
                fill opacity=0.9,
                scatter/use mapped color={
                  fill=mapped color,
                  draw=black,
                },
                ]
                \addplot3[
                  scatter,
                  scatter src=explicit,
                  mark=triangle*, % or square*, triangle*, diamond*, oplus*, x,
                ] table [meta index=3] {
                  alpha1 alpha2 alpha3 f1
                  0 0 1 86.875
                  0 0.2 0.8 86.498
                  0 0.4 0.6 86.776
                  0 0.5 0.5 86.832
                  0 0.6 0.4 68.133
                  0 0.8 0.2 68.112
                  0 1 0 68.089
                  0.01 0.01 0.98 67.955
                  0.01 0.98 0.01  87.399
                  0.2 0 0.8 87.114
                  0.2 0.2 0.6 68.033
                  0.2 0.4 0.4 86.672
                  0.2 0.6 0.2 67.923
                  0.2 0.8 0 86.893
                  0.333 0.333 0.333 86.169
                  0.4 0 0.6 84.095
                  0.4 0.2 0.4 86.343
                  0.4 0.4 0.2 68.070
                  0.4 0.6 0 86.340
                  0.5 0 0.5 87.168
                  0.5 0.5 0 74.812
                  0.6 0 0.4 86.890
                  0.6 0.2 0.2 87.165
                  0.6 0.4 0 86.057
                  0.8 0 0.2 86.970
                  0.8 0.2 0 86.574
                  0.98 0.01 0.01 87.103
                   1 0 0 86.957
                };
              \end{axis}
            \end{tikzpicture}
        % \caption{Variation of $\alpha$}
        \label{fig:alpha}
    \end{subfigure}
    \hfill
    \begin{subfigure}{0.23\textwidth}\tiny
        \centering
         \begin{tikzpicture}
              \begin{axis}[
                width=\textwidth,
                height=\textwidth,
                view={135}{35},
                xlabel={$\lambda_1$},
                ylabel={$\lambda_2$},
                zlabel={$\lambda_3$},
                xtick={0, 0.25, 0.50, 0.75, 1.0},
                ytick={0, 0.25, 0.50, 0.75, 1.0},
                ztick={0, 0.25, 0.50, 0.75, 1.0},
                xticklabel style={rotate=90, anchor=east},
                yticklabel style={rotate=45, anchor=east},
                % colormap/tab10,
                colormap/jet,
                colorbar,
                point meta min=67,
                point meta max=88,
                enlarge x limits={0.1},
                enlarge y limits={0.1},
                enlarge z limits={0.1},
                colorbar style={
                    width=0.15cm,
                    ytick={67,70,73,76,79,82,85,88},
                    yticklabel style={font=\tiny},
                    title= $\mathcal{F}_m$,
                    title style={yshift=-2ex, font=\tiny},
                  },
                only marks,
                mark=*,
                mark size=1.5pt,
                draw=black,
                line width=0.1pt,
                fill opacity=0.9,
                scatter/use mapped color={
                  fill=mapped color,
                  draw=black,
                },
                ]
                \addplot3[
                  scatter,
                  scatter src=explicit,
                  mark=triangle*, % or square*, triangle*, diamond*, oplus*, x,
                ] table [meta index=3] {
                  x     y     z     f1
                  0     0     1     82.753
                  0     0.2   0.8   85.964
                  0     0.4   0.6   75.506
                  0     0.5   0.5   68.462
                  0     0.6   0.4   75.204
                  0     0.8   0.2   85.849
                  0     1     0     78.852
                  0.2   0     0.8   85.001
                  0.2   0.2   0.6   85.105
                  0.2   0.4   0.4   68.088
                  0.2   0.6   0.2   86.938
                  0.2   0.8   0     79.386
                  0.3333 0.3333 0.3333 86.301
                  0.4   0     0.6   85.172
                  0.4   0.2   0.4   85.913
                  0.4   0.4   0.2   86.386
                  0.4   0.6   0     78.734
                  0.5   0.5   0     84.203
                  0.6   0     0.4   85.824
                  0.6   0.4   0     84.539
                  0.8   0     0.2   85.881
                  0.8   0.2   0     84.522
                  1     0     0     82.578
                  0.5   0     0.5   84.626
                  0.6   0.2   0.2   87.399
                  0.01 0.01 0.98 84.017
                  0.01 0.98 0.01 79.028
                  0.98 0.01 0.01 83.501 
                };
              \end{axis}
            \end{tikzpicture}
        % \caption{Variation of $\lambda$}
        \label{fig:lambda}
    \end{subfigure}
    % \caption{\emph{Left}: impact of $\alpha$, \emph{Right}: variation of $\lambda$}
\caption{Impact of hyperparameters $\alpha_i$ and $\lambda_i$ on HR}
    \label{fig:alpha_lambda}
\end{figure}
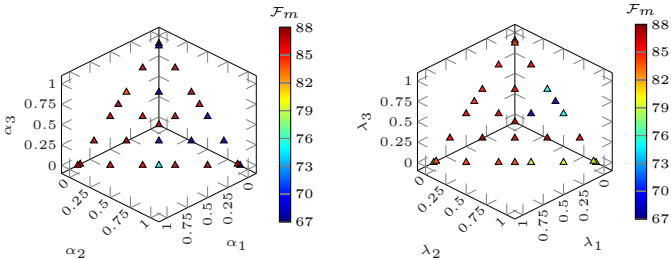

\subsection{Impact of Model Components} 

% Due to space constraints, we report the impact of the module only on the HR dataset, which is representative enough to illustrate the observed trends.

We here report the impact of modules of ProtoSiTex on HR dataset only due to space constraints, which is representative enough to illustrate the observed trends.

\subsubsection{Impact of Different Prototype Initializers} %
We evaluated four prototype initialization strategies: random, K-means centroids, Kaiming, and Xavier. As shown in Fig. \ref{fig:ablation1}(a), random initialization consistently outperformed the others across $\mathcal{F}_m$, $\mathcal{BA}_m$, and ${\mathcal{HL}}^c$. This performance gain may be attributed to the increased diversity introduced early in the training, which helps avoid premature convergence and overfitting.

% \textbf{\emph{(ii) Ablation on Alternate Training:}} 
\subsubsection{Impact of Alternate Training} 
ProtoSiTex adopts a dual-phase training strategy, alternating between unsupervised prototype learning and supervised classification. To evaluate its efficacy, we compared model variants trained with and without alternation. As shown in Fig. \ref{fig:ablation1}(b), alternate training significantly improved performance by enabling the model to discover semantically meaningful prototype clusters, and align them with downstream classification objectives. In contrast, without alternation, the model struggled to learn task-specific prototypes, lowering performance. 

\subsubsection{Impact of Similarity Function} 
We evaluated three similarity metrics: MHA, cosine similarity, and Euclidean distance, for associating subsentence embeddings with prototypes. As presented in Fig. \ref{fig:ablation1}(c), MHA of ProtoSiTex achieved superior $\mathcal{F}_m$, $\mathcal{BA}_m$, and ${\mathcal{HL}}^c$ scores by capturing richer embedding relationships, leading to more discriminative and interpretable prototype assignments.

% \begin{figure}[!t]
%     \centering
%     \scriptsize
%     (a)    \includegraphics[width=0.28\linewidth]{figures/plots/IMDb_t_SNE.png} 
%     ~(b)    \includegraphics[width=0.28\linewidth]{figures/plots/TweetEVAL_t_SNE.png}
%     ~(c)    \includegraphics[width=0.28\linewidth]{figures/plots/Ours_TSNE.png}
   
%     \caption{t-SNE plots of prototype distributions for (a) IMDb, (b) TweetEVAL, and (c) HR datasets}
%     \label{fig:tsne}
% \end{figure}

% \subsubsection{Hyperparameter Analysis:}  

\subsection{Impact of Hyperparameters}

ProtoSiTex includes several key hyperparameters: 
the number of prototypes ($q$), 
number of heads ($h$), 
loss weights for clustering ($\alpha_1$, $\alpha_2$, $\alpha_3$), and 
classification ($\lambda_1$, $\lambda_2$, $\lambda_3$). 
% We conducted controlled experiments on the Hotel Review dataset to analyze their impact.
To systematically investigate the influence of these hyperparameters on model performance, we conducted a series of experiments on the HR. 
As shown in Fig.~\ref{fig:ablation1}(d), increasing $q$ initially boosts performance by capturing diverse semantic concepts, but excessive values lead to redundancy and over-smoothing. A similar trend is observed for $h$ in Fig.~\ref{fig:ablation1}(e).  
% Furthermore, Fig.~\ref{fig:alpha_lambda} shows that assigning slightly higher weights to $\lambda_2$, $\lambda_3$, $\alpha_1$, and $\alpha_2$ yields optimal results. 
Furthermore, Fig.~\ref{fig:alpha_lambda} shows that assigning $\alpha_1=0.01$, $\alpha_2=0.98$, $\alpha_3=0.01$, and $\lambda_1=0.6$, $\lambda_2=0.2$, $\lambda_3=0.2$ yields optimal results.
This multi-level supervision prevented the model from overfitting to a single level of granularity, supporting both local and global interpretability and accuracy.

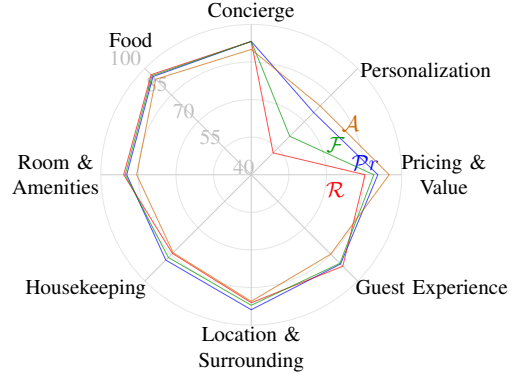
\begin{figure}[!t]
    \centering

\begin{center}
\begin{adjustbox}{scale=0.4}
\begin{tikzpicture}

% Radar grid with radial value labels along Food axis (135°)
\foreach \i/\v in {1.25/55,2.5/70,3.75/85,5/100} {
  \draw[gray!20] (0,0) circle (\i);
  \pgfmathsetmacro{\x}{-\i / 1.4142}
  \pgfmathsetmacro{\y}{\i / 1.4142}
  \node[gray!50, font=\huge, anchor=south east] at (\x,\y) {\v};
}
% Add 40 level manually (center)
\draw[gray!20] (0,0) circle (0.01);
\node[gray!50, font=\huge] at (-0.25, 0.25) {40};

% Axes
\foreach \i in {0,...,7} {
  \draw[gray!40] (0,0) -- ({360/8 * \i}:5);
}

% Axis labels
\node[align=center, font = \huge] at (6.4, 0) {Pricing \& \\ Value};
\node[align=center, font = \huge] at (5.8, 3.5) {Personalization};
\node[font = \huge] at (0, 5.4) {Concierge};
\node[align=center, font = \huge] at (-4.0, 4.5) {Food};
\node[align=center, font = \huge] at (-6.5, 0) {Room \& \\ Amenities};
\node[align=center, font = \huge] at (-5.5, -3.8) {Housekeeping};
\node[align=center, font = \huge] at (0, -5.8) {Location \& \\ Surrounding};
\node[align=center, font = \huge] at (6, -3.8) {Guest Experience};

% --- Precision polygon ---
\draw[thick, blue, opacity=0.7]
  (4.207, 0.000) -- 
  (2.072, 2.072) -- 
  (0.000, 4.435) -- 
  (-3.266, 3.266) -- 
  (-4.139, 0.000) -- 
  (-2.84, -2.84) -- 
  (0.000, -4.490) -- 
  (2.968, -2.968) -- 
  cycle;
\node[blue, font=\huge] at (3.8,0.4) {$\mathcal{P}r$};

% --- Recall polygon ---
\draw[thick, red, opacity=0.7]
  (3.783, 0.000) -- 
  (0.727, 0.727) -- 
  (0.000, 4.435) -- 
  (-3.328, 3.328) -- 
  (-4.241, 0.000) -- 
  (-2.618, -2.618) -- 
  (0.000, -4.257) -- 
  (3.042, -3.042) -- 
  cycle;
\node[red, font=\huge] at (2.8,-0.5) {$\mathcal{R}$};

% --- F1 Score polygon ---
\draw[thick, green!60!black, opacity=0.7]
  (4.072, 0.000) -- 
  (1.279, 1.279) -- 
  (0.000, 4.435) -- 
  (-3.296, 3.296) -- 
  (-4.189, 0.000) -- 
  (-2.754, -2.754) -- 
  (0.000, -4.342) -- 
  (2.946, -2.946) -- 
  cycle;
\node[green!60!black, font=\huge] at (2.8,1.0) {$\mathcal{F}$};

% --- Accuracy polygon ---
\draw[thick, orange!80!black, opacity=0.7]
  (4.584, 0.000) -- 
  (2.290, 2.290) -- 
  (0.000, 4.168) -- 
  (-3.176, 3.176) -- 
  (-3.808, 0.000) -- 
  (-2.595, -2.595) -- 
  (0.000, -4.210) -- 
  (2.641, -2.641) -- 
  cycle;
\node[orange!80!black, font=\huge] at (3.3,1.7) {$\mathcal{A}$};

\end{tikzpicture}
\end{adjustbox}
\end{center}
\caption{Aspect-wise performance of ProtoSiTex on HR}
\label{fig:aspect_radar}
\end{figure}

\begin{figure*}[!t]
    \centering
    \begin{tabular}{c|c}
        \multicolumn{2}{c}{(a)  \includegraphics[width=0.85\linewidth]{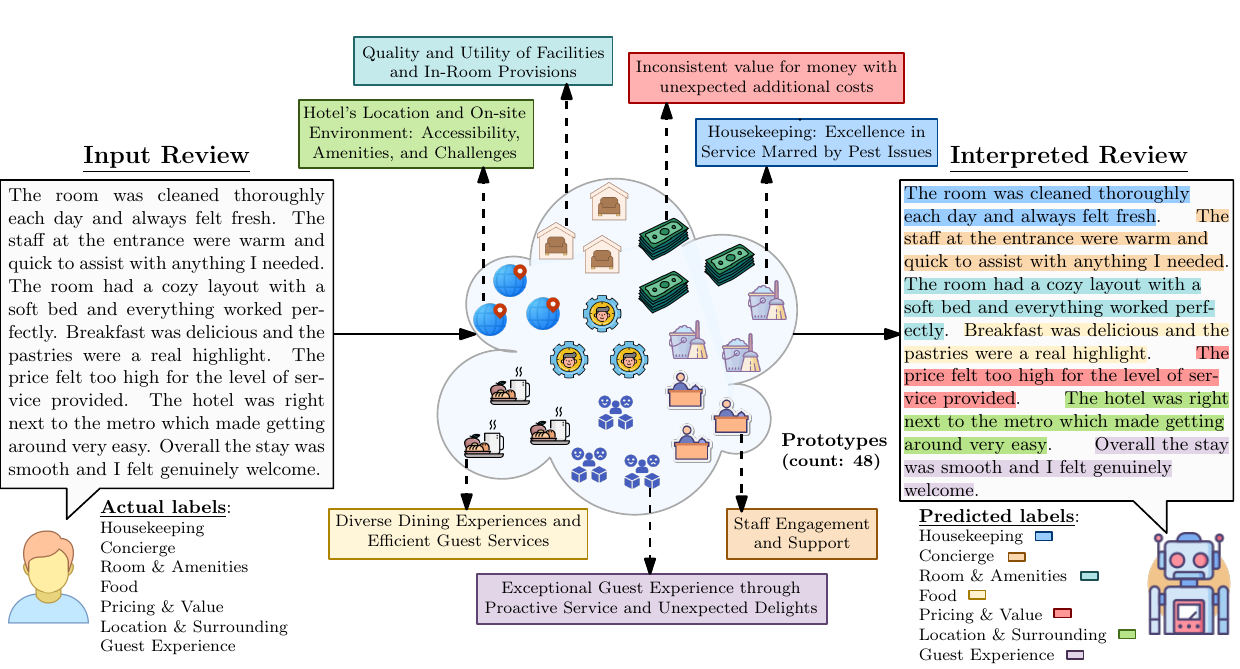}} \\ \hline 
        \\
        (b)    \includegraphics[width=0.40\linewidth]{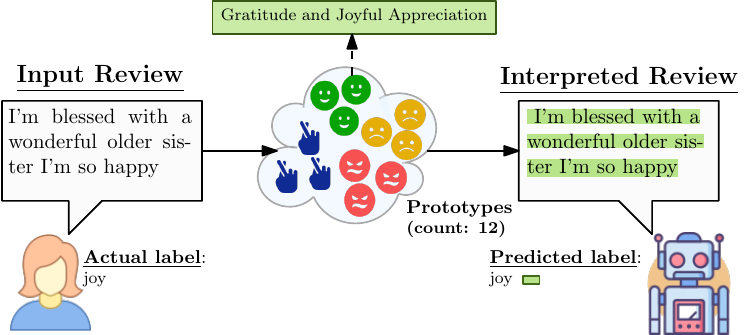} & 
        (c)    \includegraphics[width=0.47\linewidth]{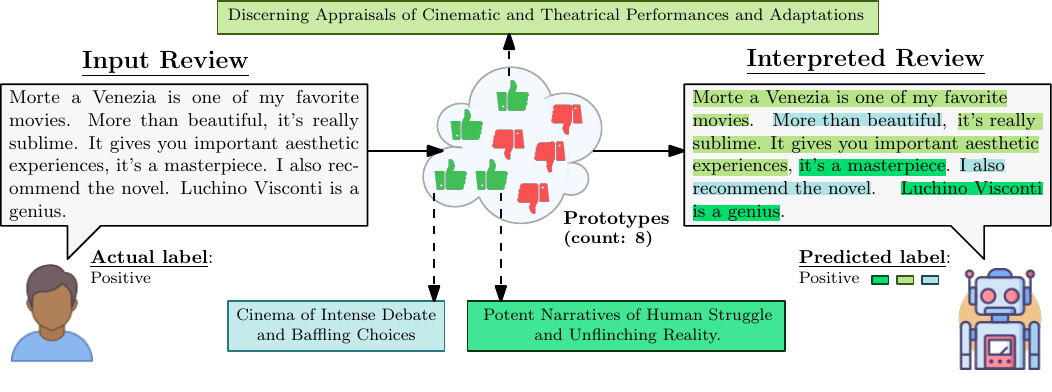}        
    \end{tabular}
    \caption{Correctly predicted examples of ProtoSiTex on (a) HR, (b) TweetEVAL \cite{TweetEVAL}, and (c) IMDb \cite{IMDb}. Best viewed in color.}
    \label{fig:qualitative_result}
\end{figure*}

\begin{figure*}[!t]
    \centering
    \begin{tabular}{c|c}
        \multicolumn{2}{c}{(a)  \includegraphics[width=0.85\linewidth]{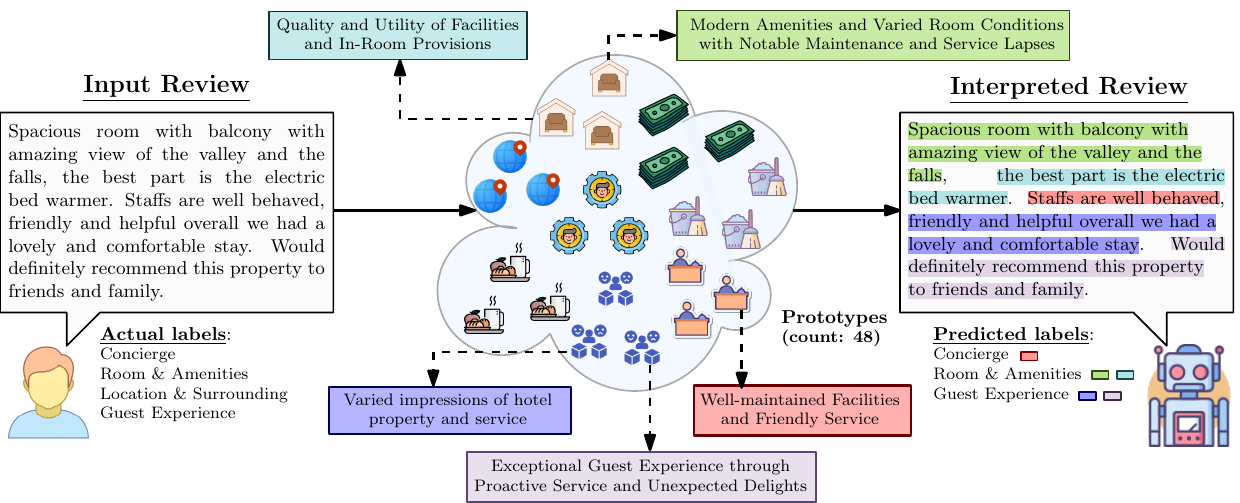}} \\ \hline
        \\
        (b)    \includegraphics[width=0.45\linewidth]{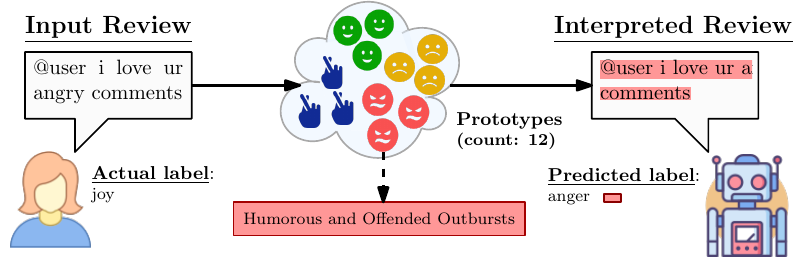} & 
        (c)    \includegraphics[width=0.45\linewidth]{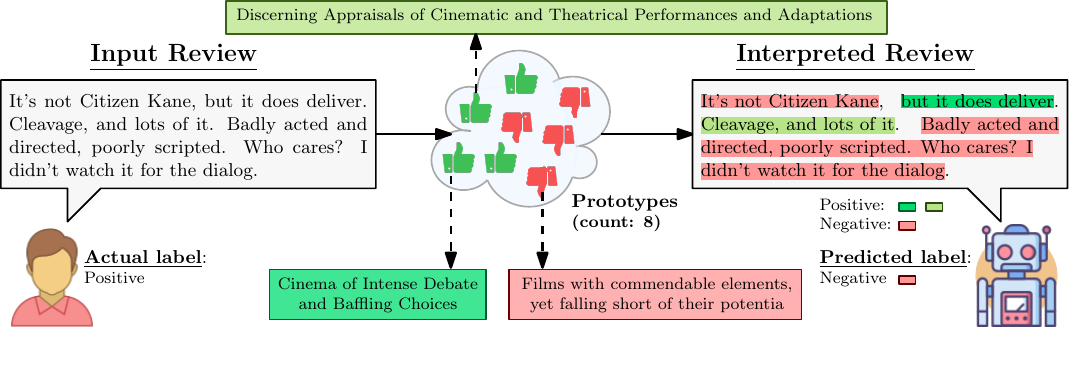}        
    \end{tabular}
    \caption{Misprediction examples of ProtoSiTex on (a) HR, (b) TweetEVAL \cite{TweetEVAL}, and (c) IMDb \cite{IMDb}. 
    Best viewed in color.}
    \label{fig:misprediction_samples}
\end{figure*}

\subsection{Aspect-wise Performance Analysis}

Fig.~\ref{fig:aspect_radar} presents a detailed aspect-wise evaluation of ProtoSiTex on the HR dataset, offering insights into the model’s performance across individual aspect categories. Notably, the aspects \emph{Guest Experience}, \emph{Food}, and \emph{Concierge} not only appear the most frequently in HR but also exhibit strong predictive performance. In contrast, \emph{Pricing \& Value} and \emph{Personalization} are the least represented. Despite the challenges posed by class imbalance, ProtoSiTex shows a strong ability to identify the \emph{Pricing \& Value} aspect, achieving $\mathcal{F}$ and $\mathcal{A}$ scores comparable to those of the more frequent categories. 
However, \emph{Personalization} remains the most challenging aspect, particularly in terms of $\mathcal{R}$, as the model often fails to capture all relevant instances. This difficulty is likely due to both its low frequency in the dataset and its inherent subjective nature. Unlike more explicit aspects such as \emph{Food} or \emph{Location \& Surrounding}, \emph{Personalization} is often expressed through subtle, indirect, and context-dependent language, making it more difficult to detect.

\subsection{Qualitative Analysis of ProtoSiTex}

To illustrate ProtoSiTex’s interpretability and prediction quality, 
% we present qualitative examples from the HR, TweetEVAL, and IMDb datasets (Fig. \ref{fig:qualitative_result}). 
we present correctly predicted examples from the HR, TweetEVAL, and IMDb datasets in Fig. \ref{fig:qualitative_result}.
In each case here, the model accurately predicts aspect or sentiment labels while grounding them in semantically coherent subsentence segments via learned prototypes. 
In the HR example, ProtoSiTex correctly identifies multi-label aspects (\emph{Room \& Amenities}, \emph{Food}, \emph{Pricing \& Value}, etc.) from a single review, while providing text-grounded justifications. This highlights its ability to disentangle overlapping review aspects and attribute them to distinct prototype-based explanations. 
In the TweetEVAL sample, ProtoSiTex captures the emotion \emph{joy} and connects it to meaningful prototypes reflecting positive sentiment, demonstrating its effectiveness in emotional reasoning with minimal text context. 
Similarly, in the IMDb example, ProtoSiTex predicts the \emph{positive} label and supports it with a rich prototype cluster centered on artistic and narrative appreciation, confirming alignment with human sentiment. 
Overall, these examples demonstrate that ProtoSiTex offers not only accurate predictions but also transparent, faithful interpretations at the subsentence level, making it suitable for real-world applications requiring interpretability.

\subsection{Misprediction Analysis}

To better understand the behavior and limitations of ProtoSiTex, we conducted a detailed analysis of mispredictions. This analysis highlights recurring patterns that lead to incorrect predictions. The most notable cases of mispredictions are discussed below:

\subsubsection{Boundary Ambiguity in Prototype} 
In multi-label classification, ProtoSiTex occasionally exhibits boundary ambiguity among prototypes, where semantically related class representations capture overlapping evidence. This ambiguity leads to confusion in distinguishing co-occurring categories, resulting in the assignment of only a single dominant label instead of multiple relevant ones. For example, in Fig. \ref{fig:misprediction_samples}(a), the review ``Spacious room with balcony with amazing view of the valley and the falls''. The actual labels are \emph{Room \& Amenities} and \emph{Location \& Surrounding}. However, ProtoSiTex predicts only \emph{Room \& Amenities}. Here, \emph{Room \& Amenities} dominates and captures shared contextual characteristics such as ``spacious room'', ``balcony'', overshadowing the equally valid cues of ``view of the valley and the falls'' that should activate \emph{Location \& Surrounding}.

\subsubsection{Lexical Cue Dominance} 
Sometimes the mispredictions in ProtoSiTex were caused by lexical bias, the tendency to rely excessively on high-intensity words when making predictions. Words such as ``angry’’, ``worst’’, or ``best’’ often dominate the model’s decision process, even when their contextual meaning is softened/ inverted by surrounding text. For instance, in Fig. \ref{fig:misprediction_samples}(b) the review ``@user i love ur angry
comments'', the actual emotion is affectionate teasing or \emph{joy}, yet the ProtoSiTex incorrectly predicts \emph{anger} due to the presence of the word ``angry''.

% {ProtoSiTex frequently failed to capture contrastive cues that shift sentiment or tone, often used to express sarcasm, humor, or a playful reversal of expectations. It introduces a contrast between two clauses, where the latter clause conveys the user’s true intent. Humans naturally recognize and interpret this transition. However, ProtoSiTex tends to anchor its prediction on the most intense local clause rather than integrating the discourse-level cues that define sarcasm and humor. (Refer Fig. \ref{fig:misprediction_samples}(c))
% }

\subsubsection{Absence of Fine-Grained Annotation} 
This observation highlights an important takeaway: even when only coarse document-level supervision is available, prototype-based explanations remain highly useful for error analysis and model trust. By showing how individual subsentences align with prototypes tied to different target classes, ProtoSiTex offers interpretable evidence that pinpoints why a misclassification occurred, rather than treating it as a black-box error. From a technical perspective, the misclassification arises due to aggregation bias, while subsentence–prototype mappings are accurate, the hierarchical aggregation process (from prototype-level activations to the final document-level decision) can overweight strong local cues and misjudge the overall label. In practice, this means even when the document-level prediction is wrong, the prototype-level reasoning still faithfully explains the decision path, helping to uncover issues such as prototype imbalance, lack of discourse-level cues, or dataset limitations tied to coarse supervision. Thus, beyond accuracy metrics, our approach demonstrates that interpretable prototype assignments provide actionable insights for debugging and model refinement. As illustrated in Fig. \ref{fig:misprediction_samples}(c), the identified prototypes, explanations, and associated class mappings are correct, even though the final document-level prediction was misclassified.

Additional experimental analyses are provided in the supplementary file. 
Appendix \textcolor{blue}{C} presents the convergence behavior of the individual and joint loss functions during training. 
Appendix \textcolor{blue}{D} offers a comprehensive evaluation of prototype interpretability using multiple metrics, including coverage, contrastivity, centered kernel alignment, and fidelity. 
Appendix \textcolor{blue}{E} reports a detailed nonparametric statistical significance analysis based on the Friedman and Nemenyi tests. 
Finally, Appendix \textcolor{blue}{F} provides extensive qualitative results, illustrating representative prototypes generated across the three benchmark datasets.

\section{Conclusion}
\label{sec:conclusion}
% \noindent
We proposed {ProtoSiTex}, a semi-interpretable framework for fine-grained, multi-label text classification. By combining adaptive prototype learning, dual-phase alternate training, and hierarchical supervision, ProtoSiTex delivers accurate and transparent predictions across multiple levels of granularity.
To support evaluation, we introduced the hotel reviews dataset (HR) with subsentence-level multi-aspect annotations. Extensive experiments on HR, along with the publicly available IMDb and TweetEVAL datasets, show that ProtoSiTex matches or exceeds baseline and state-of-the-art models, while providing faithful, human-aligned interpretations.
Future directions include extending ProtoSiTex to cross-domain and multilingual settings, incorporating multimodal cues, and integrating large language models for interactive, user-centric explanations.

\section*{Acknowledgment}
The authors heartily thank all the researchers/ interns/ annotators who helped in hotel reviews dataset preparation.

\bibliographystyle{IEEEtran}  
\bibliography{ref.bib} 

\section*{Supplementary Appendix}

Appendices A, B, C, D, E, F and G can be found in \href{https://github.com/Utsav30/ProtoSiTex_1/blob/main/Supp_Appendix_A_to_G.pdf}{https://github.com/Utsav30/ProtoSiTex1}.
\end{document}